\documentclass{article}
\usepackage{spconf,amsmath,graphicx}
\usepackage{color}

\usepackage{microtype}
\usepackage{graphicx}
\usepackage{subfigure}
\usepackage{booktabs} 
\usepackage{amsfonts}
\usepackage{breqn}
\usepackage{algorithm,algpseudocode}

\newcommand{\etal}{\textit{et al}. }
\newcommand{\ie}{\textit{i}.\textit{e}., }

\usepackage[bookmarksnumbered=true]{hyperref} 
\hypersetup{
     colorlinks = true,
     linkcolor = black,
     anchorcolor = black,
     citecolor = black,
     filecolor = black,
     urlcolor = black
     }

\title{RATE CODING OR DIRECT CODING: WHICH ONE IS BETTER  FOR \\ ACCURATE, ROBUST, AND ENERGY-EFFICIENT SPIKING NEURAL NETWORKS?}
\name{\begin{tabular}{@{}c@{}}
Youngeun Kim   \qquad
Hyoungseob Park\qquad \hspace{3mm}
Abhishek Moitra \qquad \vspace{1mm}
Abhiroop Bhattacharjee \\
Yeshwanth Venkatesha \qquad 
Priyadarshini Panda \thanks{The research was funded in part by C-BRIC, one of six centers in JUMP, a Semiconductor Research Corporation (SRC) program sponsored by DARPA, the Technology Innovation Institute (Abu Dhabi), and the National Science Foundation (Grant$\#$1947826).}
\end{tabular}}

\address{Yale University, USA}

\begin{document}
%
\maketitle

\begin{abstract}
Recent Spiking Neural Networks (SNNs) works focus on an image classification task, therefore various coding techniques have been proposed to convert an image into temporal binary spikes.
Among them, rate coding and direct coding are regarded as  prospective candidates for building a practical SNN system as they show state-of-the-art performance on large-scale datasets. 
Despite their usage, there is little attention to comparing these two coding schemes in a fair manner.
In this paper, we conduct a comprehensive analysis of the two codings from three perspectives: accuracy, adversarial robustness, and energy-efficiency. 
First, we compare the performance of two coding techniques with various  architectures and datasets.
Then, we measure the robustness of the coding techniques on two adversarial attack methods.
Finally, we compare the energy-efficiency of two coding schemes on a digital hardware platform.
Our results show that direct coding can achieve better accuracy especially for a small number of timesteps.
In contrast, rate coding shows better robustness to adversarial attacks owing to the non-differentiable spike generation process.
Rate coding also yields higher energy-efficiency than direct coding which requires multi-bit precision for the first layer.
Our study explores the characteristics of two codings, which is an important design consideration for building SNNs\footnote{The code is made available at \href{https://github.com/Intelligent-Computing-Lab-Yale/Rate-vs-Direct}{\color{blue}{https://github.com/Intelligent-Computing-Lab-Yale/Rate-vs-Direct}}}.

\end{abstract}

\section{Introduction}
\label{sec:intro}

Spiking Neural Networks (SNNs) \cite{roy2019towards,christensen20222022} have  gained increasing attention as a promising paradigm for low-power intelligence.
Inspired by biological neuronal functionality, SNNs process visual information with binary spikes over multiple timesteps.
The majority of works on SNNs have so far focused on a static image classification problem \cite{roy2019towards} to develop an energy-efficient alternative to Artificial Neural Networks (ANNs). 
Recent works utilize a backpropagation rule for training and show that this yields  state-of-the-art performance with fewer number of timesteps compared to other SNN optimization techniques
\cite{wu2019direct}.



%

To convert a static image into binary spike trains, various coding schemes have been proposed for the image classification task \cite{comsa2020temporal,park2019fast,guo2021neural}.
%
Among them, rate coding and direct coding are prominent as these coding schemes enable SNNs to be trained on large-scale datasets  \cite{wu2019direct,zheng2020going,zhang2020temporal,fang2020incorporating,lee2020enabling}.
The scalability of SNNs on large-scale datasets is important considering the growing demand for processing large-scale data on resource-constrained devices.
Rate coding converts pixel intensity into a spike train where the number of spikes is proportional to the pixel intensity \cite{diehl2015unsupervised,lee2016training,lee2020enabling}.  
On the other hand, direct coding uses a trainable layer to generate float value for each timestep \cite{wu2019direct,zheng2020going,zhang2020temporal,kim2021visual}.
%
Following the recent trend, we focus on the comparison between rate coding and direct coding even though they require more spikes than other coding techniques such as temporal coding \cite{comsa2020temporal,mostafa2017supervised} (details discussed in Section \ref{sec:relatedwork}).
In this paper, we objectively compare rate coding and direct coding in the same experimental settings from three different perspectives.
Note, it is difficult to make a fair comparison between these coding schemes from previous works \cite{wu2019direct,zheng2020going,zhang2020temporal,fang2020incorporating,lee2020enabling} which use different architectures, neuron models, and hyperparameters.
To this end, we first report the \textbf{accuracy} on various architectures (\ie multi-layer perceptron, VGG5, and VGG9) and datasets (\ie MNIST, CIFAR10, and CIFAR100).
We present the change of accuracy with respect to the number of timesteps since achieving high performance in a low-latency regime is one of the  important topics in SNN studies.
\textbf{Adversarial robustness} \cite{goodfellow2014explaining} is recently highlighted as a new feature of SNNs \cite{sharmin2020inherent}.
Kundu \etal \cite{kundu2021hire} compare the adversarial venerability of rate/direct coding and present a method to improve robustness.
We explore the adversarial robustness to provide a better understanding of the advantages and disadvantages of each coding scheme. 
Finally, \textbf{energy-efficiency} is a key evaluation metric for SNNs, therefore we estimate the energy cost of coding schemes on a digital hardware platform  \cite{chen2016eyeriss}.
%

 Our work provides several noteworthy observations for SNN design.
Direct coding yields better accuracy than  rate coding.
The advantage of direct coding increases for deeper architecture and larger datasets.
In terms of robustness, rate coding shows better robustness with respect to  Fast Gradient Sign Method (FGSM) \cite{goodfellow2014explaining} and Projected Gradient Descent (PGD) \cite{madry2017towards} attacks due to the non-differentiable Poisson spike generator.
On the other hand, since the direct coding layer is differentiable, malicious backward gradients can degrade the accuracy significantly.
Furthermore, rate coding achieves higher energy efficiency from binary input spikes than direct coding where multi-bit operation is required.

%

\begin{figure}[t]
\begin{center}
\def\arraystretch{0.5}
\begin{tabular}{@{}c@{\hskip 0.02\linewidth}c@{}c}
\includegraphics[width=0.43\linewidth]{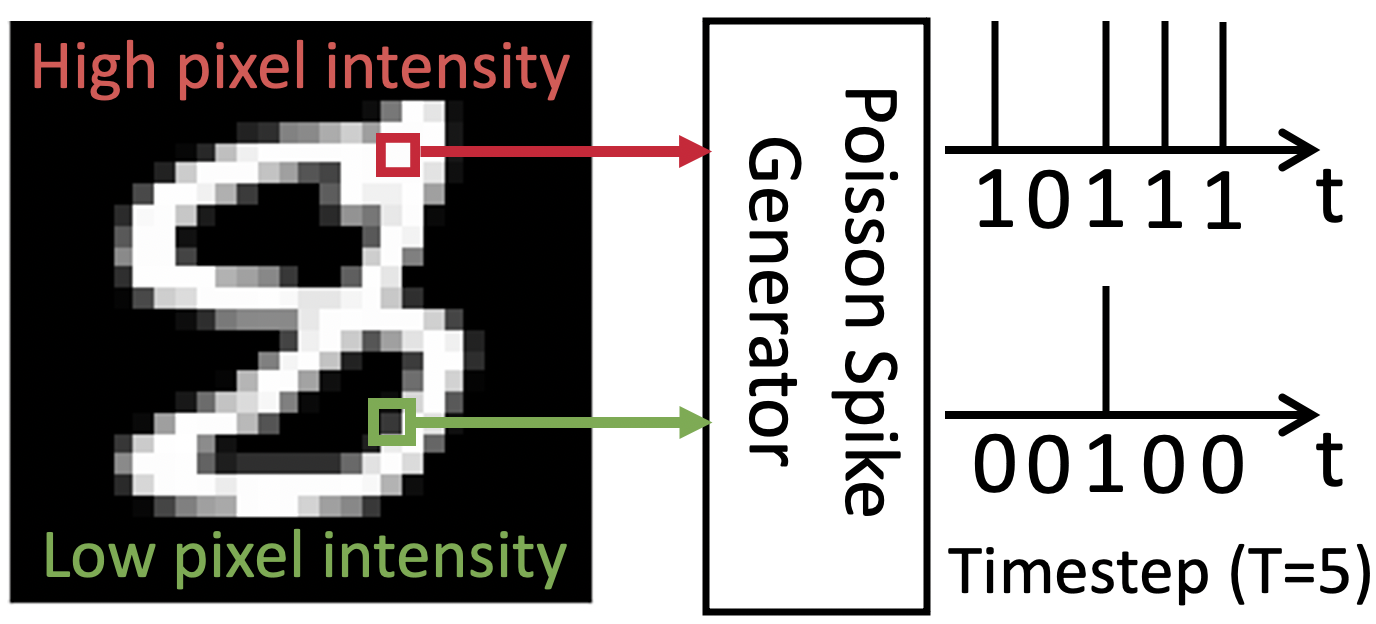} &
\includegraphics[width=0.58\linewidth]{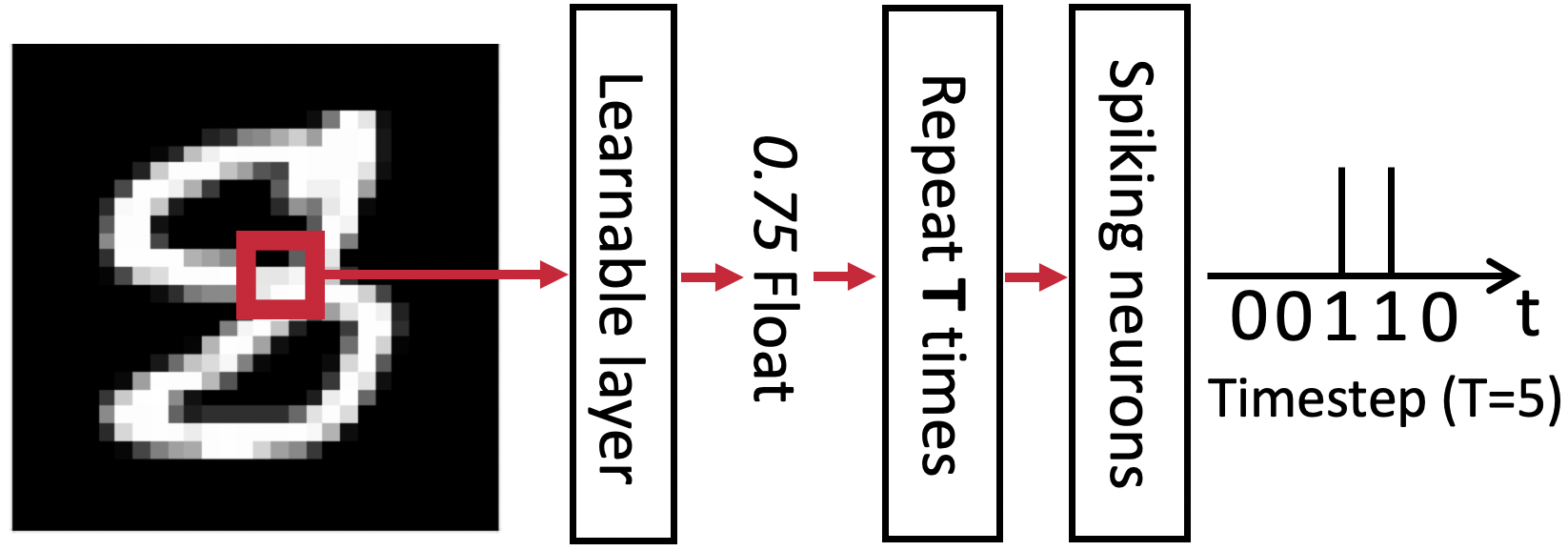} 
\\
{\hspace{1mm} (a) Rate coding } & {\hspace{0mm} (b) Direct coding }\\
\end{tabular}
\end{center}
  \vspace{-3mm}
\caption{ Illustration of rate coding and direct coding with total timesteps $T=5$.
}
  \vspace{-3mm}
\label{fig:rate_vs_direct_coding}
\end{figure}


\section{Preliminaries: Input coding schemes}
\label{sec:relatedwork}

Various coding schemes has been proposed for image classification with SNN.
Temporal coding \cite{mostafa2017supervised,comsa2020temporal} generates one spike per neuron in which spike latency is inversely proportional to the pixel intensity.
Phase coding encodes temporal information into spike patterns based on a global oscillator \cite{montemurro2008phase}.
Burst coding transmits the burst of spikes in a small-time duration, increasing the reliability of synaptic communication between neurons \cite{park2019fast}.
With these coding schemes, most prior work successfully train shallow networks, however, they are difficult to be applied when the network and dataset size are scaled up.
Rate coding can be applied on such large-scale settings, and  therefore recent state-of-the-art  methods utilize this coding \cite{diehl2015unsupervised,lee2016training,lee2020enabling}.
This coding scheme encodes the input by generating a spike train over $T$ timesteps, where the total number of spikes is proportional to the magnitude of input values. 
The spikes are sampled from a Poisson distribution. Fig. \ref{fig:rate_vs_direct_coding}(a) shows the rate coding mechanism. 
%
%
Direct coding \cite{wu2019direct} uses the floating-point inputs directly in the first layer.
As shown in Fig. \ref{fig:rate_vs_direct_coding}(b), we pass the input image (or RGB pixel values) through the first convolution layer which generates floating point outputs. Note, the float output values are repeated for $T$ time-steps of SNN processing. These outputs are then processed through a layer of spiking neurons that generates binary spikes.
Recently, Guo \etal \cite{guo2021neural} conducted a comprehensive analysis on various coding schemes.
However, they use a shallow 2-layered network with Spike-Timing-Dependent Plasticity (STDP), which is difficult to apply to deeper models working on large scale datasets.
Also, direct coding has not been compared to other coding schemes although they are leveraged in state-of-the-art works \cite{wu2019direct,zhang2020temporal}.

\vspace{-1mm}

\section{RATE coding vs. DIRECT coding}
\vspace{-2mm}
In this section, we first present the Leaky Integrate-and-Fire (LIF) neuron model and training algorithm used in our comparison.
After that, we analyze the strengths and weaknesses of two codings from three different perspectives. 

\vspace{-2mm}

\subsection{Neuron model}
\label{ssec:LIF neuron}
\vspace{-1mm}
We use LIF neuron for our comparison.
The membrane potential $U_m$ of LIF neuron stores the temporal spike information. 
When an input signal $I(t)$ is given to the LIF neuron, the membrane potential is varied: 
\begin{equation}
    \tau_m \frac{dU_m}{dt} = -U_m  + I(t),
    \label{eq:LIF_origin}
\end{equation}
where,  $\tau_m$ is the time constant for the membrane potential decay.
Since the voltage and current have continuous values, we convert the differential equation into a discrete version following the previous works \cite{wu2019direct,fang2020incorporating}.
For each timestep $t$, we can formulate the membrane potential $u_{i}^{t}$ of a single neuron $i$ as: 
\vspace{-0.5mm}
\begin{equation}
    u_i^t = (1 - \frac{1}{\tau_m})  u_i^{t-1} + \frac{1}{\tau_m} \sum_j w_{ij}o^t_j.
    \label{eq:LIF}
\vspace{-1mm}
\end{equation}
Here, the current membrane potential consists of the decayed membrane potential from previous timesteps and the weighted spike signal from the pre-synaptic neurons.
The notation  $w_{ij}$ is for weight connections between neuron $i$ and neuron $j$.
The neuron $i$ accumulates voltage and generates a spike output $o_i^{t}$ whenever $u_i^{t}$ exceeds the firing threshold $\theta$.
After the neuron fires, the membrane potential is lowered by the amount of the threshold.

\begin{figure*}[t]
\begin{center}
\def\arraystretch{0.5}
\begin{tabular}{@{}c@{\hskip 0.01\linewidth}c@{}c@{}c@{\hskip 0.01\linewidth}c@{}}
\includegraphics[width=0.25\linewidth]{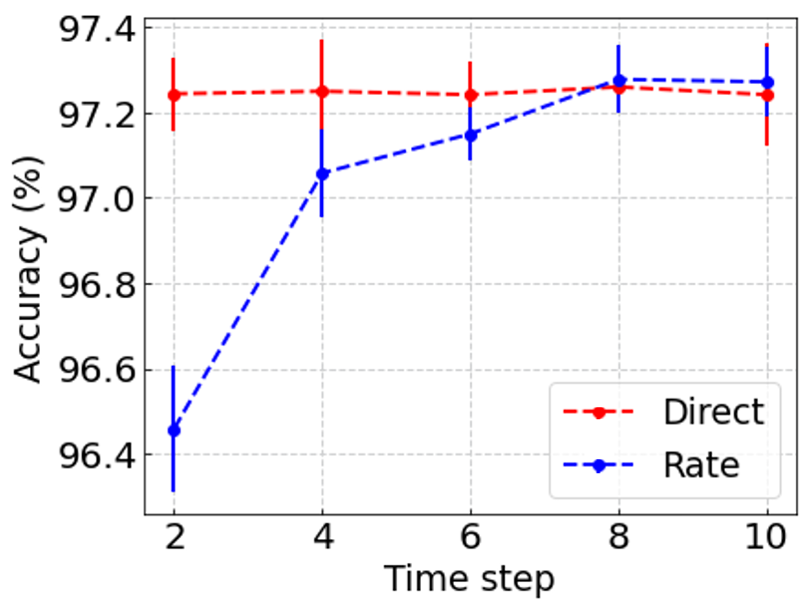} &
\includegraphics[width=0.25\linewidth]{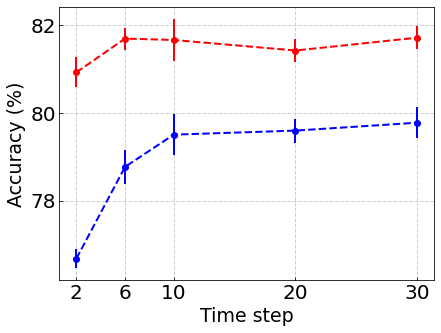} &
\includegraphics[width=0.25\linewidth]{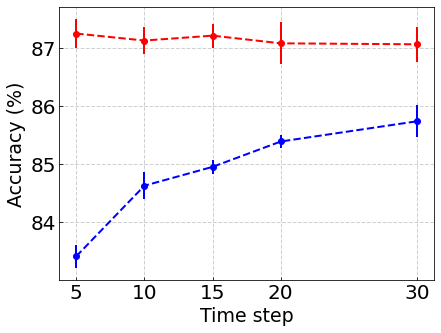} &
\includegraphics[width=0.25\linewidth]{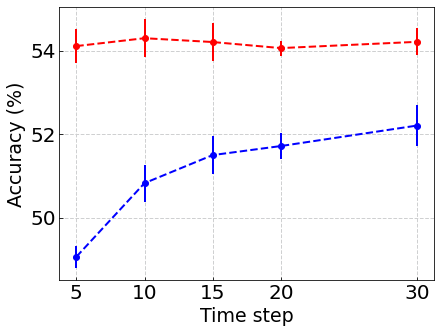}
\\
{\hspace{3mm} (a) MLP-MNIST } & {\hspace{3mm} (b) VGG5-CIFAR10  } & {\hspace{5mm} (c) VGG9-CIFAR10 } & 
{\hspace{4mm} (d) VGG9-CIFAR100  }\\
\end{tabular}
\end{center}
  \vspace{-6mm}
\caption{The accuracy change with respect to the number of timesteps. We run each configuration 5 times and report the mean and standard deviation (vertical line). 
}
  \vspace{-1mm}
\label{fig:comparison}
\end{figure*}

\begin{figure*}[t]
\begin{center}
\def\arraystretch{0.5}
\begin{tabular}{@{}c@{\hskip 0.01\linewidth}c@{}c@{}c@{\hskip 0.01\linewidth}c@{}}
\includegraphics[width=0.25\linewidth]{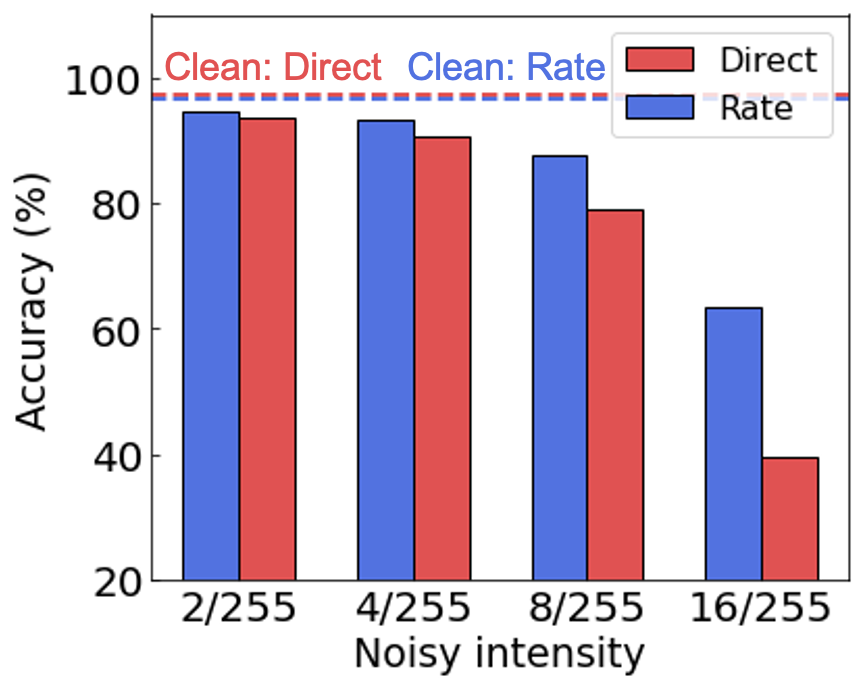} &
\includegraphics[width=0.25\linewidth]{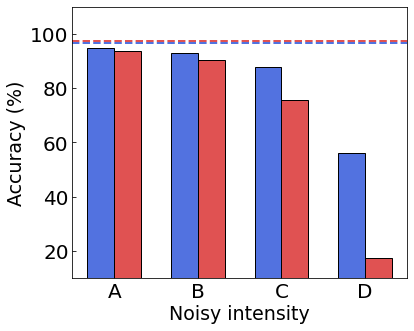} &
\includegraphics[width=0.25\linewidth]{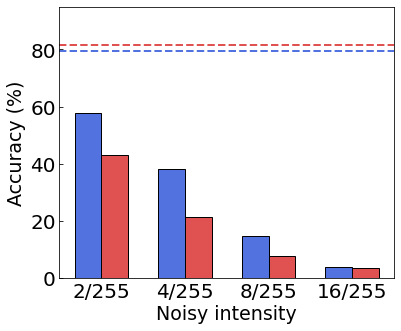} &
\includegraphics[width=0.25\linewidth]{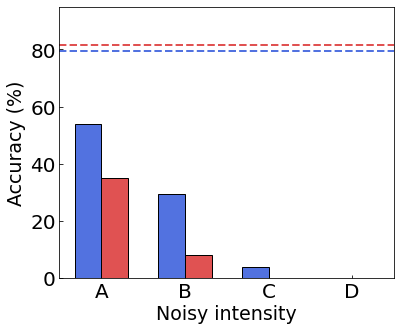}
\\
{\hspace{5mm} (a) MLP-FGSM } & {\hspace{5mm} (b) MLP-PGD  } & {\hspace{2mm} (c) VGG5-FGSM } & 
{\hspace{2mm} (d) VGG5-PGD   }\\
\end{tabular}
\end{center}
  \vspace{-6mm}
\caption{The adversarial robustness with respect to FGSM attack and PGD attack.
We present the clean accuracy using dotted lines.
In the PGD attack experiments, A-PGD[2/255, 1/255,10],  B-PGD[4/255, 1/255,10], C-PGD[8/255, 4/255,10], D-PGD[16/255, 4/255,10]. We use timestep 10 for all experiments.
}
  \vspace{-1mm}
\label{fig:noise_intensity}
\end{figure*}

\vspace{-2mm}

\subsection{Optimizing SNN using backpropagation}

\vspace{-1mm}

Various optimization techniques for SNNs have been proposed in the past decade.
%
Among them,  surrogate gradient learning
has been mainly utilized in recent SNN works 
 because it shows state-of-the-art performance in a low-latency regime \cite{zheng2020going,zhang2020temporal,fang2020incorporating,kim2020revisiting,kim2021optimizing}.
Further, surrogate gradient based backpropagation learning can be implemented using well-established machine learning frameworks like PyTorch \cite{paszke2017automatic}.
In our comparison, we use surrogate gradient learning as a baseline optimization technique.

Given input spikes, we train SNNs based on gradient optimization. 
Intermediate LIF neurons accumulate pre-synaptic spikes and generate output spikes (Eq. \ref{eq:LIF}).
Spike information is passed through all layers and stacked or accumulated at the output layer (\ie prediction layer).
This enables the accumulated temporal spikes to be represented as probability distribution after the softmax function.
From the accumulated membrane potential, we can compute the cross-entropy loss for SNNs.
Based on the calculated loss, we compute the gradients of each layer $l$.
Here, we use spatio-temporal back-propagation (STBP), which accumulates the gradients over all timesteps \cite{wu2018spatio}.
We can formulate the gradients at the layer $l$ by chain rule as:
\begin{equation}
      \frac{\partial L}{\partial W_l} =
\begin{cases}
 \sum_{t}(\frac{\partial L}{\partial O_l^t}\frac{\partial O_l^t}{\partial U_l^t} + \frac{\partial L}{\partial U_l^{t+1}}  \frac{\partial U_l^{t+1}}{\partial U_l^{t}})
 \frac{\partial U_l^t}{\partial W_l},  & \text{if $l:$ hidden } \\
    \frac{\partial L}{\partial U_l^T}\frac{\partial U_l^T}{\partial W_l}.
    & \text{if $l:$ output} 
\end{cases}
\label{eq:delta_W}
\end{equation}
Here, $O^t_l$ and $U^t_l$ are output spikes and membrane potential at timestep $t$ for layer $l$, respectively.
LIF neurons in hidden layers  generate  spike output only if the membrane potential $u_i^t$ exceeds the firing threshold, leading to non-differentiability.
To deal with this problem, we use an approximate gradient:
\vspace{-1mm}
\begin{equation}
    \frac{\partial o_i^t}{\partial u_i^t} = \max \{0, 1-  \ | \frac{u_i^t - \theta}{\theta} \ | \}.
    \label{eq:approx_grad_function}
\end{equation}
%
Overall,  network parameters at the layer $l$ are updated  based on the gradient value (Eq. \ref{eq:delta_W}).
\vspace{-1mm}

\subsection{Experimental settings}

\vspace{-1mm}

In our experiments, we use three architectures  on (\ie MLP 784-800-10, VGG5, and VGG9)  three public datasets (\ie MNIST, CIFAR10, and CIFAR100).
{MNIST} contains gray-scale images of size 28 $\times$ 28. 
{CIFAR10} consists of 60,000 RGB color images of size 32 $\times$ 32.  (50,000 for training / 10,000 for testing) with 10 categories. 
{CIFAR100} has the same configuration as CIFAR10, except it contains 100 categories.
For all datasets, we use random horizontal flip for data augmentation.
Our implementation is based on PyTorch framework \cite{paszke2017automatic}. 
We set the total number of epochs to 60, 100, and 100 for MNIST, CIFAR10, and CIFAR100, respectively.
During training, we utilize step-wise learning rate scheduling with a decay factor of 10 at 50\% and 75\% of the total epochs. 
We train the networks with Adam optimizer with an initial learning rate $1e-4$.
For LIF neuron, we set time constant $\tau_m$ and threshold $\theta$ to $2$ and $1$, respectively.


\vspace{-3mm}

\subsection{Accuracy comparison}
\vspace{-2mm}
Fig. \ref{fig:comparison} shows the accuracy of rate-encoded SNN and direct-encoded SNN with respect to the number of timesteps.
From the experimental results, we observe the following:
(1) In general, direct coding brings higher accuracy than rate coding especially with small number of timesteps.
(2) As the rate-coded SNN is trained with larger number of timesteps, the performance gap between the two coding decreases.
(3) As the dataset and network architecture gets more complicated, the performance gap between the two coding increases. 
 The reason for higher performance of  direct-coded SNNs is a trainable input coding layer (Fig. \ref{fig:rate_vs_direct_coding}(b)). 
The weights of the coding layer are trained for minimizing cross-entropy loss, providing optimal float amplitude.

\vspace{-2mm}

\subsection{Robustness studies}
\vspace{-2mm}
We compare the robustness of rate-coded and direct-coded SNNs against adversarial attacks. %
For this work, we consider two adversarial attacks.
FGSM attack \cite{goodfellow2014explaining} is a single-step attack based on backward gradients.
For the input image $x$,  adversarial image $x_{adv}$ is generated by adding the sign of gradients scaled by $\epsilon$:
\vspace{-1mm}
    \begin{equation}
    x_{adv} = x + \epsilon ~sign(\nabla_x(\mathcal{L}(x,y_{true}))),
    \label{fgsm}
    \end{equation}
where, $y_{true}$ stands for the ground-truth label.
{Projected Gradient Descent (PGD)}  attack \cite{madry2017towards} is an iterative adversarial attack characterized by parameters, such as, maximum perturbation $\epsilon$, perturbation step size $\alpha$ and the number of iterations $n$. 
Higher values of $\epsilon$, $\alpha$ and $n$ denote stronger PGD attacks. 
In our experiments, we represent the configuration of PGD as [$\epsilon$, $\alpha$, $n$].
%
In Fig. \ref{fig:noise_intensity}, we compare the robustness of SNNs with rate coding and direct coding on two architectures (\ie MLP and VGG5). 
Note, for generating adversarial examples for SNNs with rate coding, we use the method proposed by \cite{sharmin2020inherent}. Interestingly, we find that rate-coded SNNs have higher adversarial robustness approximately up to 20\% higher accuracy compared to direct-coded SNNs.  
This is because the Poisson generator function is non-differentiable, therefore  the gradient  has to be approximated, making the attack ineffective.




\subsection{Energy-efficiency of two coding schemes}

\begin{figure}[t!]
    \centering
    \includegraphics[width=0.88 \linewidth]{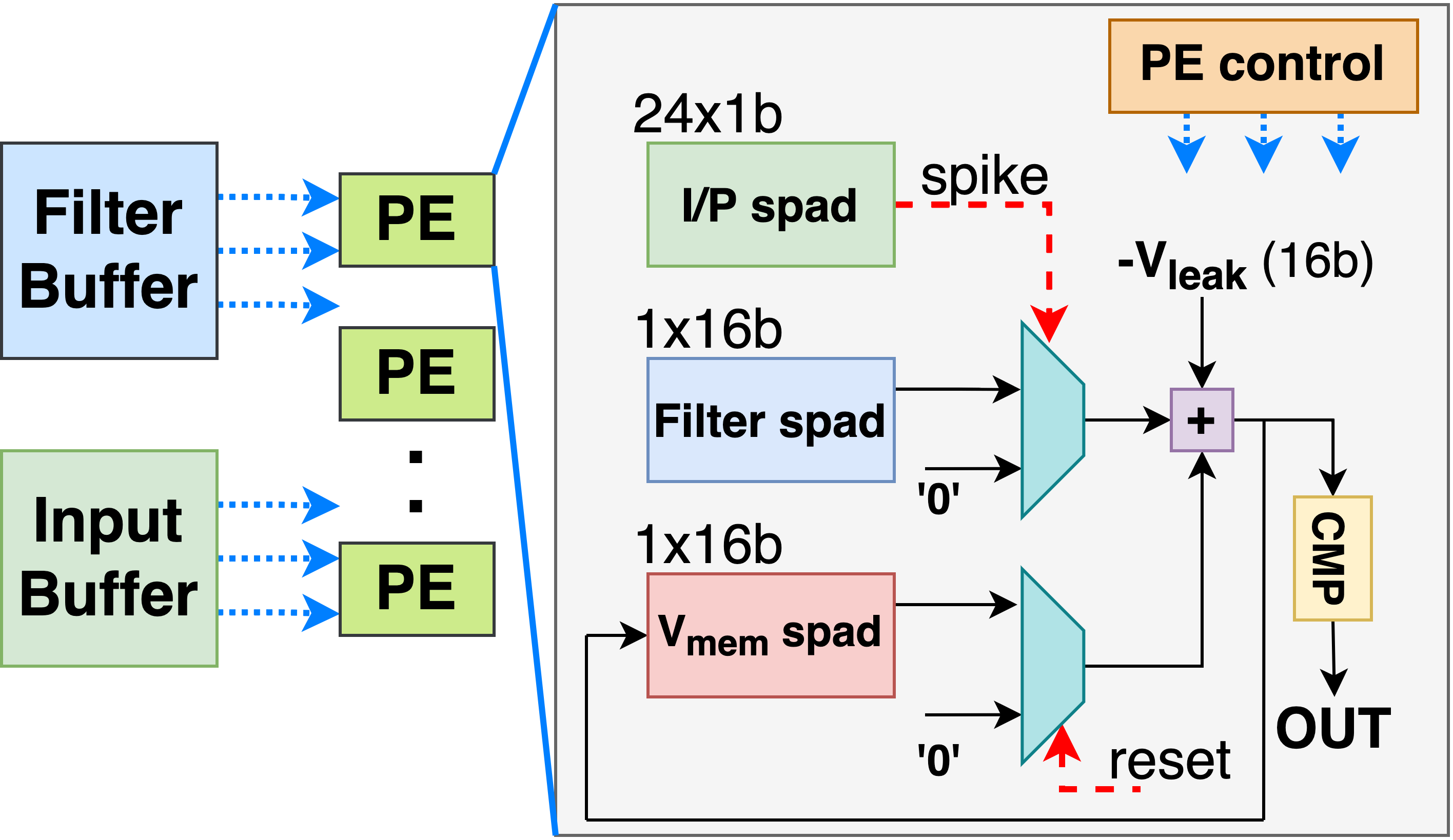}%
    \vspace{-1mm}
    \caption{A pictorial representation of a PE in a 16-bit \textit{Eyeriss} platform for SNN evaluation.}
    \label{eyeriss_pe}
        \vspace{-2mm}
\end{figure}

To assess the energy costs incurred by the two coding schemes, we evaluate our MLP and VGG5 SNNs on a 16-bit \textit{Eyeriss} platform (65 nm CMOS technology) using an \textit{output stationary} dataflow. \textit{Eyeriss} follows a von-Neumann mode of neural computation widely adopted in modern accelerators \cite{chen2016eyeriss, narayanan2020spinalflow}. It enables us to optimize over different design choices such as, type of dataflow, computation reuse and skipping zero computations. A representation of a processing element (PE) for the evaluation of SNNs on \textit{Eyeriss} has been depicted in Fig. \ref{eyeriss_pe}. 
Rate-encoded SNNs also involve a digital circuit for achieving the Poisson rate coding that is also accounted for in our energy evaluation \cite{guo2021neural}.
Table \ref{tab:energy_val} shows the normalized energy values per image for the SNNs, wherein we find that rate coding yields less energy cost than direct coding by $\sim 50\%$. Here, all the measured energy values are normalized with respect to a 16-bit Multiply-and-Accumulate (MAC) operation. Primarily, it is due to the fact that the first layer of the direct-encoded SNNs has 16-bit data precision and thus, incurs greater computational costs with respect to the rate-encoded SNNs having binary spike trains as inputs. 
Note, the first layer mainly contributes to the total energy cost as deep layers yield highly sparse spike activation.
Moreover, in the standard \textit{Eyeriss} architecture, direct coding requires the same full-precision inputs and weights for the first layer to be repeatedly fetched $T$ times to perform corresponding MAC operations (where, $T$ is total number of timesteps) that add significantly to the energy expenditure. On the other hand, rate-encoded SNNs with a sparse distribution of input spikes help reduce the computations and hence, energy expenditure on the \textit{Eyeriss} platform. In order to circumvent the repetition of the full-precision first layer computations in direct-encoded SNNs, we need to modify the standard \textit{Eyeriss} PE to facilitate digital shift operations for the partial sums generated during MAC. The modification will alleviate repeated fetches of the same inputs and weights from the spads (or registers) as well as the cost of multiply operation for multiple time-steps. 

\begin{table}[t !]
\centering
\caption{Energy expended by rate-encoded and direct-encoded SNNs on an \textit{Eyeriss} platform normalized with respect to the energy of 1 MAC operation. We use 10 timesteps for both architectures.}
\vspace{4mm}
\label{tab:energy_val}
\resizebox{0.8\linewidth}{!}{%
\begin{tabular}{|c|c|c|}
\hline
\textbf{} & \multicolumn{2}{c|}{Normalized energy/image} \\ \hline
Scheme         & MLP/MNIST  & VGG5/CIFAR10  \\ \hline
Rate coding   & 8.58E+06           & 2.73E+07              \\ \hline
Direct coding & 2.26E+07           & 4.22E+07              \\ \hline
\end{tabular}%
}
\end{table}


\vspace{-2mm}

\section{CONCLUSION and DISCUSSION}
\vspace{-2mm}
This study is motivated by the question: \textit{Which coding scheme is better for building accurate, robust, and energy-efficient SNN, rate coding or direct coding?}
Note that we focus on rate coding and direct coding because they enable SNNs to be trained on large-scale datasets with deep architectures.
This question is timely as neuromorphic researchers are finding a way of scaling up the SNN models like their ANN counterparts.
To explore this, we conduct a comprehensive analysis on accuracy, robustness, and energy-efficiency.
Direct coding trains the weights of the coding layer, therefore the input image can be converted into the optimal spikes with float amplitude for minimizing cross-entropy loss. 
Consequently, direct coding achieves better accuracy than rate coding where the input spikes are generated based on pixel intensity.
Although the learnable layer in direct coding improves performance, such layer is vulnerable to adversarial attack as it conveys malicious backward gradients to the input image.
On the other hand, Poisson spike generator of the rate coding is non-differentiable and stochastic, resulting in less performance degradation in case of adversarial attack.
We also find that rate-encoded SNNs achieve higher energy-efficiency on
\textit{Eyeriss} architecture than their direct-encoded counterparts.
 Overall, our extensive experiments provide a fundamental understanding of the two coding schemes.
We hope this enables researchers to utilize proper coding schemes according to their specific applications.

\newpage

\bibliographystyle{IEEEbib}
\bibliography{refs}

\end{document}